\pdfoutput=1

\documentclass[11pt]{article}

\usepackage[final]{acl}

\usepackage{times}
\usepackage{latexsym}
\usepackage{algorithm2e}
\usepackage{todonotes}

\usepackage[T1]{fontenc}
\usepackage{amssymb}
\usepackage{amsmath}
\usepackage[utf8]{inputenc}
\usepackage{amsmath}
\usepackage{algorithm}
\usepackage{microtype}

\usepackage{inconsolata}
\usepackage{colortbl}
\usepackage{graphicx}
\usepackage{microtype}
\usepackage{graphicx}
\usepackage{subfigure}
\usepackage{booktabs} 
\usepackage{tcolorbox}
\usepackage{listings}
\usepackage{fancyvrb}
\usepackage{lipsum}

\usepackage{hyperref}

\title{Hypothesis Generation for Materials Discovery and Design Using Goal-Driven and Constraint-Guided LLM Agents}

\author{Shrinidhi Kumbhar \quad Venkatesh Mishra \quad Kevin Coutinho \quad  Divij Handa \\  \textbf{Ashif Iquebal} \quad  \textbf{Chitta Baral} \\
  Arizona State University\\
\small{\texttt{\{skumbha4, vmishr23, chitta\}@asu.edu}}
}

\begin{document}
\maketitle
\begin{abstract}

Materials discovery and design are essential for advancing technology across various industries by enabling the development of application-specific materials. Recent research has leveraged Large Language Models (LLMs) to accelerate this process. We explore the potential of LLMs to generate viable hypotheses that, once validated, can expedite materials discovery. Collaborating with materials science experts, we curated a novel dataset from recent journal publications, featuring real-world goals, constraints, and methods for designing real-world applications. Using this dataset, we test LLM-based agents that generate hypotheses for achieving given goals under specific constraints. To assess the relevance and quality of these hypotheses, we propose a novel scalable evaluation metric that emulates the process a materials scientist would use to evaluate a hypothesis critically. Our curated dataset, proposed method, and evaluation framework aim to advance future research in accelerating materials discovery and design with LLMs. \footnote{Data and code are available at \href{https://github.com/shri071/Hypothesis-Generation-for-Materials-Discovery-and-Design-Using-Goal-Driven-and-Constraint-Guided-LLM}{https://github.com/shri071/Hypothesis-Generation-for-Materials-Discovery-and-Design-Using-Goal-Driven-and-Constraint-Guided-LLM}}

\end{abstract}

\section{Introduction}
\label{submission}

\begin{figure}[h!]
    \centering
    \includegraphics[width=0.95\columnwidth]{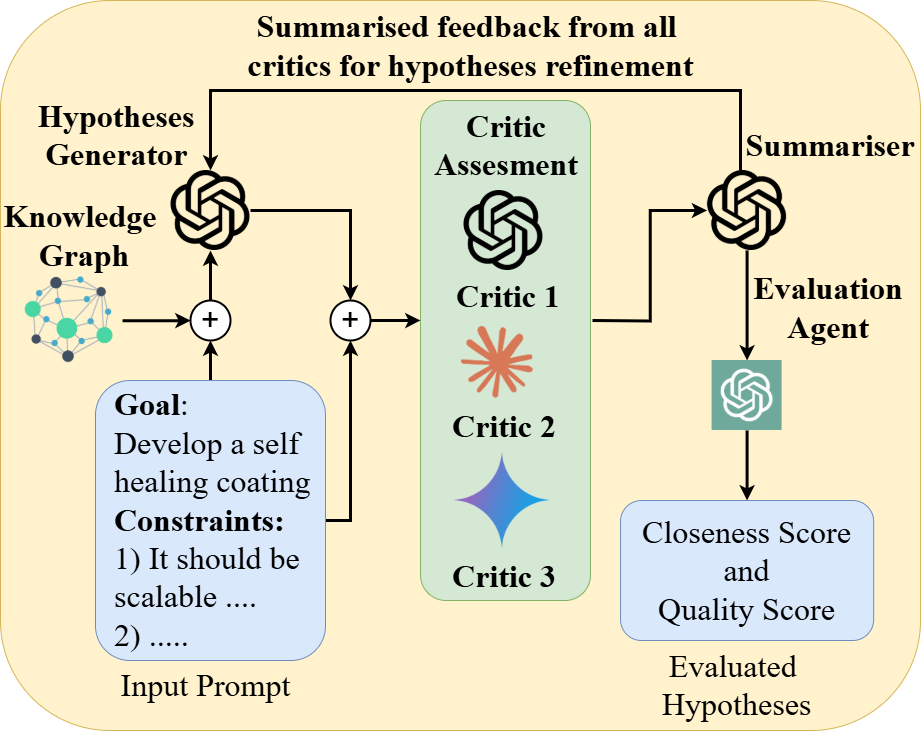}
    \caption{Overview of our iterative hypothesis generation and evaluation pipeline. Starting from an input prompt and a knowledge graph, the Hypotheses Generator (GPT-4o) proposes 20 hypotheses, which are then reviewed by three critics--GPT-4o, Claude-3.5-Sonnet, and Gemini-1.5-Flash. Their feedback is consolidated by the Summarizer (GPT-4o); if unanimous agreement is not reached, the hypotheses along with critic feedback are fed back to the Hypotheses Generator for refinement and are re-evaluated by the critics. Once approved, the final hypotheses proceed to the Evaluation Agent (OpenAI-o1-preview) for scoring.}
    \label{fig:hypothesis-generation-framework}
\end{figure}

The discovery and design of materials to meet specific application needs is essential to address critical challenges we face today \citep{Jain2013CommentaryTM}. Traditional methods for materials discovery and design are time-intensive and resource-heavy, requiring researchers to conduct extensive literature reviews, explore vast compositional, chemical, and structural spaces through simulations, and perform laborious lab-based experiments \citep{davies2016computational,hautier2012computer}. Recent advances in machine learning and data-driven approaches have accelerated materials discovery by enabling predictions of material structures and properties \citep{liu2017materials,oganov2019structure} and in proposing novel materials \citep{chen2022universal, merchant2023scaling, ren2022invertible, fung2022atomic}. However, these methods rely on extensive training datasets and can't process natural language, limiting their flexibility for hypothesis generation.

To address these limitations, recent research in natural language processing (NLP) and materials science has explored the use of LLMs for hypothesis generation \citep{jia2024llmatdesign,sprueill2024chemreasoner,ghafarollahi2024sciagents}. While promising, these methods are often restricted to a specific material or property and rely on domain-specific external tools that run simulations, which are costly and time-intensive. To overcome these challenges, we design an LLM-based agent for \textbf{ACCEL}erating \textbf{MAT}erials discovery and design, \textbf{\textsc{AccelMat}}. Specifically, our architecture consists of a Hypotheses Generation Agent, a multi-LLM Critic system with iterative feedback, a Summarizer Agent to consolidate all feedback, and an Evaluation Agent to assess the hypotheses. An overview of our architecture is shown in Fig \ref{fig:hypothesis-generation-framework}.

To assess the performance of our system, we introduce \textbf{\textsc{MatDesign}}, a dataset developed in collaboration with materials science experts. While existing benchmarks, such as those proposed by \citet{zaki2023mascqa} and \citet{guo2023can}, have proven valuable for evaluating LLMs knowledge and capability in material science and chemistry tasks, they are limited to assessing LLMs knowledge within graduate-level subdomains of materials science or narrowly focused chemistry tasks, failing to evaluate their capability to generate hypotheses for real-world materials discovery and design tailored for a specific application under given constraints. Furthermore, our dataset is constructed from research papers published in leading journals in 2024, ensuring it lies beyond the knowledge cutoff of all LLMs employed in our study\footnote{Llama-3.1-70B (\href{https://github.com/meta-llama/llama3/blob/main/MODEL_CARD.md}{Model Card}) and GPT-4o (\href{https://platform.openai.com/docs/models\#gpt-4o}{Documentation}) have knowledge cutoffs of December 2023 and October 2023 respectively.}.

Finally, to evaluate the hypotheses generated by our agent, the evaluation metric is divided into two primary components: Closeness and Quality. Closeness measures how close the generated hypothesis is with the ground truth. Quality assesses the Alignment, Scientific Plausibility, Novelty, Feasibility, Scalability, Testability and Impact Potential of the hypothesis within its domain. These metrics mirror the systematic and rigorous approach employed by material scientists when validating hypotheses, providing a robust framework for comprehensive evaluation.



In summary, our contributions are as follows:
\begin{itemize}
    \item Create a novel benchmark, \textsc{\textbf{MatDesign}}, consisting of goals and constraints, along with their corresponding materials and methods.
    \item Develop an LLM-based agentic framework to generate and refine material discovery and design hypotheses.
    \item Propose a scalable evaluation metric to measure the relevance and quality of the generated hypotheses for material discovery and design.
\end{itemize}

\begin{table}[ht]
\centering
\resizebox{\columnwidth}{!}{
\begin{tabular}{c|c|c|c} 
\toprule
LLM Agent & Diverse Mat & Diverse Prop & {Tool Free} \\ \midrule
 LLMatDesign       & \checkmark            & \texttimes             & \checkmark         \\ \midrule
 ChemReasoner      & \texttimes            & \checkmark             & \texttimes         \\ \midrule
 SciAgents         & \texttimes            & \checkmark             & \checkmark         \\ \midrule
 \rowcolor[RGB]{200, 230, 255}\textbf{\textsc{AccelMat}} & \checkmark            & \checkmark             & \checkmark         \\
\bottomrule
\end{tabular}
}
\caption{Comparison of existing LLM agent frameworks with our framework, \textbf{\textsc{AccelMat}}, for materials design and discovery hypotheses generation. Our framework covers a wide range of materials and properties while being independent of domain-specific tools.}
\label{table: agent-framework-comparison}
\end{table}

\begin{table}[ht]
\centering
\resizebox{\columnwidth}{!}{
\begin{tabular}{c|c|c|>{\columncolor[RGB]{200,230,255}}c} 
\toprule
Dataset              & MaScQA      & ChemLLMBench & \textbf{\textsc{MatDesign}} \\
\midrule
 Real World Constr.  & \texttimes  & \texttimes   & \checkmark         \\ \midrule
 Mat. Design Prob.   & \texttimes  & \texttimes   & \checkmark         \\ \midrule
 No Data Leakage     & \texttimes  & \texttimes   & \checkmark         \\ \midrule
 Difficulty Level    & Graduate   & Research     & Research           \\
\bottomrule
\end{tabular}
}
\caption{Comparison of existing benchmarks to ours--\textbf{\textsc{MatDesign}}. Our benchmark provides real-world constraints and goals. It is not present in the pre-training of LLMs and contains more complex problems.}
\label{table: dataset-comparison}
\end{table}

\section{Background and Related Work}

\subsection{LLMs for Materials Discovery and Design}
Machine learning and data-driven methods have significantly accelerated materials science research by reducing the time and resources required for discovery and design \citep{liu2017materials, oganov2019structure}. Generative models have furthered these advancements, as evidenced by works like \citet{chen2022universal, merchant2023scaling, court20203, xie2021crystal, ren2022invertible, fung2022atomic, long2021constrained}, which have provided researchers with powerful tools to aid material discovery and design. Recently, LLMs have emerged as a promising approach in this domain, with efforts falling into two main categories.

The first category involves fine-tuning LLMs on domain-specific datasets or corpora \citep{sirumalla24multi, ozccelik2024chemical, song2023honeybee, Jacobs2024Regression}. By training the models on materials science data, researchers can enhance their ability to predict material properties or address domain-specific questions. However, these approaches are resource-intensive and rely heavily on the availability of extensive, high-quality datasets, which can be challenging to compile.

The second category focuses on leveraging agent-based frameworks that integrate LLMs with domain-specific tools, APIs, and databases to facilitate hypothesis generation and refinement \citep{jia2024llmatdesign, sprueill2024chemreasoner, ghafarollahi2024sciagents, zhang2024honeycomb}. These multi-agent systems enable iterative improvements by combining LLM outputs with specialized domain knowledge, simulations, and experimental feedback. While effective, these methods are often constrained by the scope of available tools, the specificity of the materials or properties studied, and the accessibility of supporting resources. This limits their applicability and ease of use across broader material science challenges.

To address these limitations, our work introduces an LLM-based agentic framework designed to support hypothesis generation and exploration across a wide range of material science applications. As summarized in Table \ref{table: agent-framework-comparison}, unlike existing methods, our approach minimizes dependency on domain-specific tools and enhances generalizability, enabling more accessible and scalable materials discovery and design workflows. Parallel efforts by \citet{ding2024matexpert} and \citet{yang2024moose}, have explored distinct approaches, with the former leveraging databases to identify materials with desired properties and utilizing fine-tuned LLMs to suggest modifications, and the latter employing literature-based insights and mutation algorithms to rediscover chemical hypotheses.

\subsection{Benchmarks for Materials Discovery and Design with LLMs} 
The increasing use of LLMs in scientific domains \citep{choi2024accelerating, taylor2022galactica, chen2022scientific, cavanagh2024smileyllama, singhal2022large} underscores the importance of rigorous evaluation frameworks to assess their performance. Existing benchmarks, such as those by \citet{guo2023can} and \citet{zaki2023mascqa} assess LLMs on a range of chemistry and material science tasks respectively. However, these benchmarks fail to assess LLM's ability to generate valid hypotheses for materials discovery or design under specific goals and constraints. Furthermore, these benchmarks are susceptible to overlap with LLM pretraining corpora, particularly given that most open-source and closed-source LLMs have a knowledge cutoff in late 2023. This overlap raises concerns about the novelty and validity of benchmark evaluation.

To mitigate these issues, we developed a new dataset in collaboration with materials science experts. This dataset, derived from 50 research papers published in journals from January 2024, includes entries structured with clearly defined goal statements, relevant constraints, and the materials and methods required to achieve the specified objectives. As summarized in Table \ref{table: dataset-comparison}, our benchmark not only ensures independence from pretraining data but also emphasizes real-world applicability by focusing on hypothesis generation tasks, providing a robust foundation for evaluating LLMs in materials discovery and design.

\section{\textsc{MatDesign}}

\begin{table*}[ht]
\resizebox{\textwidth}{!}{
\begin{tabular}{p{5cm}|p{6cm}|p{6cm}} 
\toprule
\multicolumn{1}{c|}{\textbf{Goal}} & \multicolumn{1}{c|}{\textbf{Constraints}} & \multicolumn{1}{c}{\textbf{Material Name and Methods}} \\
\midrule
Develop a scalable extrinsic self-healing coating system for corrosion protection of metallic structures in offshore environments. &
1) The material should incorporate a self-healing mechanism triggered by a single environmental factor (e.g., water). \newline
2) The self-healing material should allow multiple healing events. \newline
\newline \vdots &
\textbf{Materials}: Core-shell nanofibers synthesized using coaxial electrospinning. Organosilane compounds, specifically silyl esters, used as the self-healing agent. Metallic substrates (e.g., steel) for corrosion tests. \newline
\textbf{Methods}: Coaxial electrospinning of core-shell nanofibers with an organosilane compound (silyl ester) as the healing agent.
\newline \vdots
\\ 
\bottomrule
\end{tabular}
}
\caption{An instance from our dataset, \textsc{\textbf{MatDesign}}, extracted from the paper by \citet{spera2024self}. The Goal and Constraints are provided as the input to our designed framework. Material Name and Methods are used as ground truth for evaluation.}
\label{table: dataset-instance}
\end{table*}


To evaluate the capacity of LLMs to generate materials discovery and design hypotheses, we curated a specialized benchmark. The dataset consists of information extracted from 50 research papers published from January 2024 in prominent journals such as Nature, Nature Communications, and Progress in Organic Coatings. Each entry in the dataset includes the following manually extracted and structured components:
\begin{itemize}
\item \textbf{Goal Statement}: Description of material development for a specific application, serving as the first part of the input to our framework.
\item \textbf{Constraints}: Conditions that steer the hypothesis generation process, serving as the final part of the input to the framework.
\item \textbf{Materials}: Key materials and their compositions, serving as the first part of the ground truth.
\item \textbf{Methods}: Brief description of synthesis methods, forming the final part of the ground truth.
\end{itemize}

Materials Science experts assisted in all extractions, ensuring the benchmark's accuracy and relevance. The selection of publications from January 2024 is essential to ensure that the ground truth information (material names, compositions, and synthesis methods) associated with the respective goals and constraints is not present in the training corpus of LLMs we use, which have a training cutoff of late 2023. This temporal constraint enables a rigorous evaluation of the LLMs ability to generate genuinely novel hypotheses, relying on their internalized understanding of materials science principles rather than retrieving pre-existing information. An example instance is shown in Table \ref{table: dataset-instance}.




\section{\textsc{AccelMat}}
The LLM-based multi-agent framework designed for this study consists of four key components, each playing a critical role in the iterative process of hypothesis generation and refinement. Since our aim is to aid material scientists by generating novel and viable hypotheses, we use state-of-the-art proprietary LLMs. Performance using open-source LLMs is provided in Appendix \ref{appendix:open-source}.

\begin{enumerate}
    \item \textbf{Hypotheses Generation Agent (HGA)}: Given the Goal Statement and Constraints, the HGA generates multiple hypotheses, accompanied by reasoning for each. This agent is powered by GPT-4o. The prompts used for HGA are provided in Appendix \ref{box: hypothesis-generation-agent}.
    \item \textbf{Critic Agents (CA)}: The second component consists of three Critic Agents--GPT-4o, Claude-3.5-Sonnet \citet{anthropic_claude_sonnet}, and Gemini-1.5-Flash \citet{team2024gemini}. These agents are provided with the hypotheses generated by the HGA, goal statement, and constraints. Their role is to evaluate each hypothesis, assessing its alignment with the goal and constraints. Each critic gives detailed feedback to guide subsequent hypothesis refinement cycles. The prompts used for CA can be found in Appendix \ref{box: critic-agent}.
    \item \textbf{Summarizer Agent (SA)}: The Summarizer Agent consolidates and organizes the feedback from all three CAs into a structured format. It then provides this comprehensive feedback to the HGA to guide the refinement process. We use GPT-4o as the SA. The prompt used for SA is provided in Appendix \ref{box: summarizer-agent}.
    \item \textbf{Evaluation Agent (EA)}: The Evaluation Agent is used to evaluate the closeness and quality of the generated hypotheses. We use OpenAI-o1-preview \citet{jaech2024openai} as our evaluation agent.
\end{enumerate}
The HGA, CA, and SA form the hypothesis generation and refinement framework. The EA is used solely to evaluate the generated hypotheses.

\section{Evaluation Metrics}
\label{section:evaluation}
To assess the generated hypotheses comprehensively, we adopt a dual-metric evaluation framework. The first metric, Closeness, measures the degree of alignment between the generated and ground truth hypotheses. The second metric, Quality, evaluates the generated hypotheses based on six distinct criteria. Together, these metrics provide a holistic assessment: Closeness facilitates evaluation in scenarios with ground truth data, while Quality enables robust evaluation in cases where ground truth hypotheses are unavailable. The evaluation prompts used for measuring Closeness and Quality can be found in Appendix \ref{Evaluation Prompts for OpenAI-o1-preview}. 

\subsection{Closeness}
The Closeness metric measures the similarity between the generated and ground truth hypotheses in the dataset, focusing on the following:
\begin{enumerate}
    \item \textbf{Concept Overlap:} Assesses the degree to which the core ideas, methods, and scientific concepts in the generated hypothesis align with those in the ground truth.
    
    \item \textbf{Property Overlap:} Evaluates the extent to which the material properties in the generated hypothesis align with those in the ground truth, encompassing both quantitative values and qualitative descriptions.
    
    \item \textbf{Keyword Matching:} Assesses the accuracy of specific entities (e.g., material names, chemical compounds, synthesis methods) and keywords in the generated hypothesis by comparing them to those in the ground truth.
\end{enumerate}
The objective scores for evaluating Closeness are presented in Table \ref{table: closeness-eval}.

\begin{table*}[ht]
\centering
\resizebox{\textwidth}{!}{
\begin{tabular}{p{3cm}|p{3cm}|p{3cm}|p{3cm}|p{3cm}|p{3cm}} 
\toprule
\textbf{Evaluation Metric} & \textbf{Score 1} & \textbf{Score 2} & \textbf{Score 3} & \textbf{Score 4} & \textbf{Score 5} \\
\midrule
\textbf{Concept Overlap} & No Overlap: The generated hypothesis contains entirely different concepts. & Minimal Overlap: A few general ideas may be similar, but most key concepts differ. & Moderate Overlap: Some core ideas overlap, but critical concepts are missing or misrepresented. & High Overlap: Most core concepts match with minor differences. & Complete Overlap: The generated hypothesis fully mirrors the core ideas of the ground truth. \\
\midrule
\textbf{Property Overlap} & Not Similar: No overlapping or similar properties to the ground truth. & Slightly Similar: A few properties match, but most are different in magnitude or type. & Moderately Similar: Some important properties are similar, though others differ. & Highly Similar: Most key properties align well with the ground truth. & Perfect Match: The generated hypothesis fully matches the properties of the ground truth. \\
\midrule
\textbf{Keyword Matching} & No Match: None of the keywords or entities match. & Minimal Match: A small number of non-critical keywords match. & Partial Match: Several important keywords and entities match, but some are missing. & High Match: Most critical keywords and entities match, with minor discrepancies. & Complete Match: All key keywords and entities match exactly with the ground truth. \\
\bottomrule
\end{tabular}
}
\caption{Evaluation Metrics Scores for Closeness between Generated Hypotheses and Ground Truth}
\label{table: closeness-eval}
\end{table*}

\subsection{Quality}
The second part of the evaluation focuses on the Quality of the generated hypotheses, assessed across six defined criteria developed in collaboration with materials science experts. This metric is particularly valuable for evaluating hypotheses in scenarios where no ground truth is available for the given goals and constraints. The detailed definition associated with the ratings is described in Table \ref{table: six-criteria-for-eval}. The definition of each criterion is provided below:

\begin{enumerate}
    \item \textbf{Alignment with Research Objectives and Constraints:} Evaluates how effectively the generated hypothesis addresses the specified objectives in the goal statement while satisfying all provided constraints.

    \item \textbf{Scientific Plausibility:} Evaluates whether the generated hypothesis is consistent with established scientific principles and theories.

    \item \textbf{Innovation \& Novelty:} Measures the extent to which the generated hypothesis presents original ideas, approaches, or methods.

    \item \textbf{Testability:} Evaluates how easily and effectively the generated hypothesis can be tested experimentally, with the available techniques, equipment, and resources.

    \item \textbf{Feasibility \& Scalability:} Assesses the practicality of implementing the generated hypothesis across different scales, from laboratory experiments to industrial applications, considering factors such as existing infrastructure, cost, and the effort required for scaling.

    \item \textbf{Impact Potential:} Assesses the generated hypothesis's potential to significantly advance the field or address critical challenges.
\end{enumerate}

The overall Quality score is calculated as the unweighted average of the scores across all six criteria. While a weighted average approach could be used to emphasize specific criteria, the unweighted average is used in this study for simplicity.

To enable scalable and accurate evaluation, we utilize OpenAI-o1-preview. This model was selected over traditional metrics like ROUGE, which are limited in capturing nuanced semantic similarities. While embedding-based models offer an alternative, they often require fine-tuning on domain-specific datasets and may fail to adequately capture contextual meanings critical for evaluating hypotheses in materials science. For comparative purposes, traditional evaluation is provided in Appendix \ref{Closeness Evaluation with BERTScore}. A set of generated hypotheses evaluated with the Evaluation Agent and our proposed evaluation metric can be found in Appendix \ref{Evaluated Hypotheses by OpenAI-o1-preview}.

\small
\begin{algorithm}
\small
    \SetAlgoLined
    \SetKw{Initialize}{Initialize}
    \textbf{Require:} Goal Statement \(G_S\), Constraint List \(C\), Knowledge Graph \(K_G\), Critic LLMs: [\(C_1\), \(C_2\), \(C_3\)] \\
    Hypotheses Generator \(H_G\), Summarizer \(S\) \\
    Number of Cycles \(max\_cycles = 5\)\\
    \(H_H \leftarrow [ ]\) \tcp{Hypothesis History initialized as empty list}
    \(F_{AC} \leftarrow [ ]\) \tcp{Feedback from all critics initialized as empty list}
    \textbf{Generate Initial Hypothesis:}\\
    \(V_h \gets H_G(G_S, C, K_G)\)\\
    \(H_H.append(V_h)\)\\
    \If{all \(V_h\) agreed by \(C_1\), \(C_2\), \(C_3\)}{
        \Return \(V_h\)\\
    }
    \Else{
        \For{\(cycle = 1\) to \(max\_cycles\)}{
            \(F_{AC} \leftarrow feedback(G_S, C, H_H\left[ -1 \right])\) \tcp{Feedback from all critics}
            \(S_F \leftarrow summariser(F_{AC})\) \tcp{Summarized Feedback from all critics}
            \(R_h \leftarrow H_G(G_S, C, K_G, H_H\left[ -1 \right], S_F)\) \tcp{Refined Hypothesis}
            \(H_H.append(R_h)\)\\
            \(A_S \leftarrow intersection(F_{AC})\) \tcp{Agreed Suggestions}
            \If{\(len(A_S) == 20\)}{
                \textbf{break}
            }
        }
    }
    \Return \(H_H\left[ -1 \right]\)
    
    \caption{Hypotheses Generation with Critic Feedback and Knowledge Graphs}
\end{algorithm}
\label{fig:algorithm}
\normalsize


\section{Experiments}
\label{Experiemnts}
We evaluate the performance of our multi-LLM agent, \textsc{\textbf{AccelMat}}, on the proposed benchmark, \textsc{\textbf{MatDesign}} using the evaluation methodology outlined in Section \ref{section:evaluation}. Drawing inspiration from concepts and agent configurations employed in diverse domains \citet{ghafarollahi2024sciagents, ouyang2023structured, gao2024empowering, jia2024llmatdesign}, we designed three distinct configurations of \textsc{\textbf{AccelMat}} tailored for material discovery and design, with the first configuration as a baseline mentioned in the section \ref{Hypotheses Generation without Feedback from Critics_method}. These configurations are detailed in the subsections below.

\subsection{Hypotheses Generation without Feedback from Critics}
\label{Hypotheses Generation without Feedback from Critics_method}

As a baseline, for this configuration, we evaluate the standalone hypothesis generation capabilities of the HGA without external feedback. The HGA generates 20 hypotheses with detailed reasoning based on a given goal statement and constraints. These hypotheses are independently reviewed by three CAs, who assess their alignment with the goal and adherence to constraints. Only hypotheses unanimously validated by all three CAs are finalized for evaluation, ensuring reliability through consensus. This setup isolates the HGA’s performance, focusing on its ability to generate high-quality hypotheses autonomously. Refer to Appendix \ref{box: hypothesis-generation-without-feedback} for the prompts used for this configuration.

\subsection{Hypotheses Generation with Feedback from Critics}

This configuration introduces an iterative feedback loop involving the HGA, three CAs, and SA to improve the Closeness and Quality of the hypotheses. The HGA generates 20 hypotheses, which the CAs evaluate for alignment with a goal and constraints. If unanimous agreement is not reached, each CA provides individual feedback, which the SA consolidates into cohesive feedback for the HGA to refine its hypotheses. This process iterates up to five times or stops early if consensus is achieved, finalizing unanimously validated hypotheses for evaluation. Refer to Appendix \ref{box: hypothesis-generation-with-feedback-from-critics} for prompts for all agents used in this configuration.


\subsection{Hypotheses Generation with Knowledge Graph and Feedback from Critics}

This configuration builds upon the second setup by incorporating a comprehensive materials science knowledge graph. In addition to the HGA, CAs, and SA components, this configuration integrates contextual information from MatKG \citep{venugopal2024matkg}, the largest publicly available knowledge graph in materials science.

To tailor the hypothesis generation process, keywords relevant to specific applications mentioned are extracted from the goal statement and constraints using GPT-4o. These keywords are then utilized to query MatKG, retrieving relevant materials and their associated properties. The extracted information is subsequently provided to the HGA as supplementary context, enabling it to generate hypotheses that are both novel and aligned with the predefined goals and constraints.

By leveraging the extensive and structured data within MatKG, this setup enhances the HGA's ability to propose innovative and efficient material combinations. Detailed prompts for this configuration can be found in Appendix \ref{box: hypothesis-generation-with-KG-and-feedback}.

\section{Results and Analysis}

\begin{figure*}[h!]
    \centering
    \includegraphics[width=\textwidth]{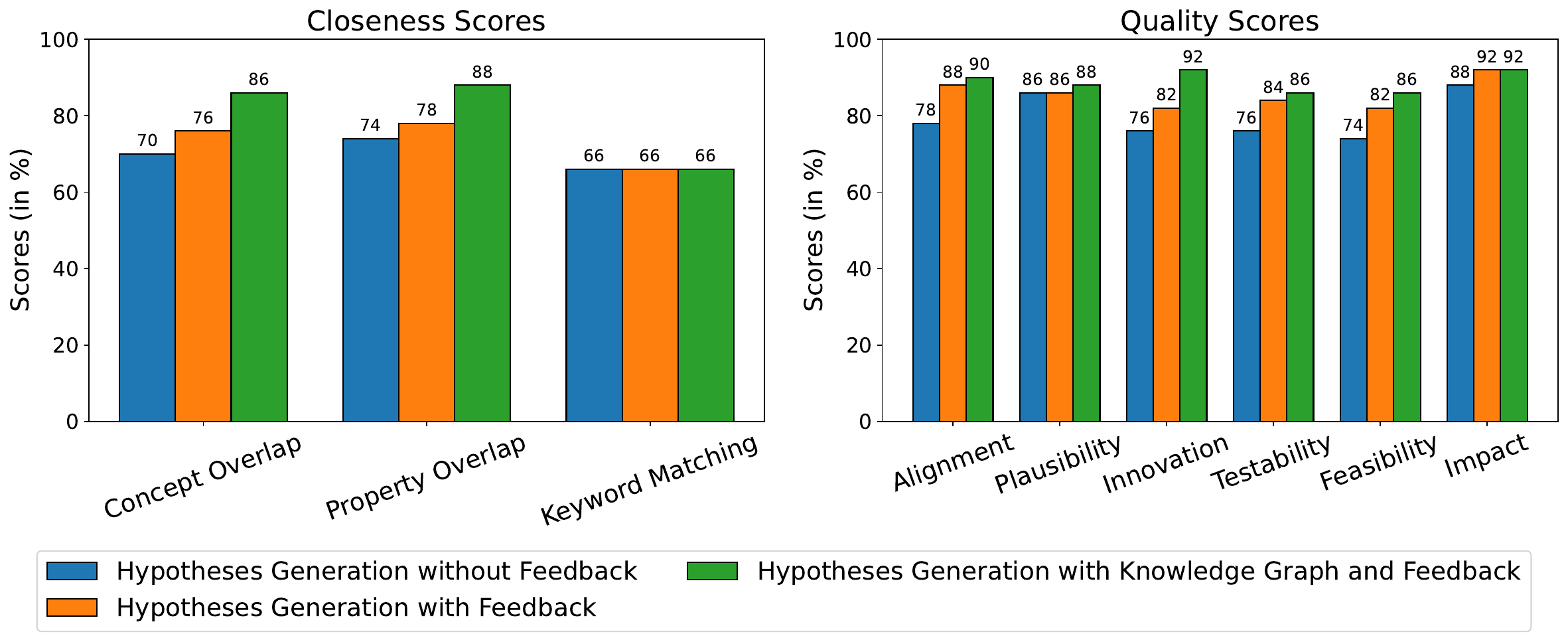}
    \caption{The left plot illustrates the Closeness metric scores across three evaluation criteria for the three configurations. The right plot depicts the Quality metric scores across six evaluation criteria for the same configurations. Both plots highlight that integrating feedback from Critic Agents and leveraging contextual knowledge from the Knowledge Graph enhances performance.}
    \label{fig:results-close-source}
\end{figure*}

This section presents the analysis of the performance of various configurations of the \textsc{\textbf{AccelMat}}. The results are summarized in Figure \ref{fig:results-close-source}.

\paragraph{Hypotheses Generation without Feedback}
The first configuration examined is the HGA operating without critical feedback loops. This baseline configuration revealed several challenges:

\begin{itemize}
    \item \textbf{Lack of Consensus:} Generated hypotheses frequently failed to achieve unanimous agreement among critics, indicating inconsistencies in the reasoning and alignment with the provided goal.
    \item \textbf{Incomplete Adherence to Constraints:} A significant number of hypotheses did not fully respect the constraints set for material selection. Missing key details in the reasoning process was a recurring issue.
    \item \textbf{Bias in Material and Method Selection:} The system exhibited a noticeable preference for certain materials and methods, which constrained the exploration of alternative solutions and reduced diversity in the output.
\end{itemize}

These shortcomings resulted in hypotheses that were often too generic and lacked the detail needed for practical implementation. This can be seen from the lowest feasibility score.

Quantitatively, the average closeness score for this configuration was 70\%, which was 10\% lower than the best-performing configuration. Similarly, the Quality metric suffered, achieving only 79.67\%, representing a 9.33\% decline relative to the optimal setup. These results underscore the limitations of operating without a feedback mechanism.

\paragraph{Hypotheses Generation with Feedback}
Introducing feedback loops from critics into the HGA demonstrated substantial improvements across all metrics:

\begin{itemize}
    \item \textbf{Enhanced Constraint Adherence:} Feedback iterations enabled hypotheses to align more closely with predefined goals and constraints, as visible with increased Alignment score and Closeness score.
    \item \textbf{Increased Diversity and Feasibility:} Feedback facilitated exploration of a broader range of materials and methods, reflecting improved conceptual understanding and hypothesis diversity. The diverse suggestions were equally feasible, demonstrating a balance between variety and practicality in addressing the given goals and constraints.
    \item \textbf{Refined Methodology:} Successive iterations improved the procedural reasoning and robustness of generated hypotheses, resulting in more actionable outputs.
\end{itemize}

This feedback-enabled configuration achieved a Closeness score of 73.33\%, a 3.33\% improvement, and a Quality score of 85.67\%, marking a 6\% increase over the feedback-free setup.

However, limitations persisted. The hypotheses remained focused on well-established methodologies, with limited exploration of unconventional solutions. While the reasoning for material selection improved, it still lacked sufficient depth to propose unique yet actionable alternatives. These results emphasize the importance of feedback loops in enhancing system performance while pointing to the need for additional mechanisms to promote innovation.

\paragraph{Hypotheses Generation with Knowledge Graph and Feedback}
Integrating a knowledge graph and feedback from critics yielded the best performance, demonstrating significant advantages:

\begin{itemize}
    \item \textbf{Diverse and Novel Combinations:} The use of the knowledge graph enabled the exploration of more diverse and novel material combinations, facilitating innovative approaches to material selection and design. This resulted in hypotheses that were both unique and well-aligned with the constraints and objectives.
    \item \textbf{Consensus Among Critics:} The integration of feedback ensured a highest rate of agreement among critics, indicating a strong alignment with design goals and constraints.
    \item \textbf{Improved Feasibility:} The majority of the hypotheses generated were not only diverse but also highly feasible, reflecting the robustness of this configuration.
\end{itemize}

Quantitatively, this configuration achieved the highest Closeness score of 80\%, a 6.67\% improvement over the feedback-only setup, and a Quality score of 89\%, marking a 3.33\% increase.

Despite these successes, in some cases, the information retrieved from the knowledge graph, while 
relevant to the goal and constraints, tends to focus on fundamental materials. This resulted in broad 
and generalized suggestions that lack the specificity needed for practical implementation. These findings demonstrate the effectiveness of combining knowledge graphs with feedback while highlighting opportunities for refining specificity and depth.

\paragraph{Increased Consensus among Critics with Feedback and Knowledge Graph}
Generating more hypotheses increases the likelihood of aligning with the ground truth by improving the chances of capturing correct outputs that can assist material science researchers. However, we wanted to restrict the cost per instance and the runtime, hence we restricted ACCELMAT to generate 20 hypotheses per goal even though we agree that more hypotheses would potentially be better. As shown in Table \ref{tab:hypothesis_agreement}, incorporating critic feedback and grounded knowledge from MatKG increased consensus among critics, with an average 19 out of 20 generated suggestions reaching agreement after consecutive feedback cycles. This demonstrates the effectiveness of the critic feedback system in maximizing viable hypothesis generation.

\begin{table}[ht]
\centering
\resizebox{\columnwidth}{!}{
\begin{tabular}{c|c} 
\toprule
ACCELMAT Framework Type                & Avg No. of Hypotheses \\ & Agreed by Critics / 20 \\
\midrule
 Without FB             & 11  \\ \midrule
 With FB              & 18  \\ \midrule
 \rowcolor[RGB]{200,230,255} \textbf{With FB and KG}       & \textbf{19}  \\
\bottomrule
\end{tabular}
}
\caption{Comparison of the average number of hypotheses agreed upon by all critics (out of 20) across different LLM-based agentic frameworks. Incorporating knowledge graphs and critic feedback maximizes the generation of viable hypotheses.}
\label{tab:hypothesis_agreement}
\end{table}

\subsection{Human Expert Evaluation}
A set of 42 suggestions, encompassing all three configurations, were independently evaluated by four PhD students in Materials Science. Their assessments employed the same evaluation metrics used by the automated system.

For the first configuration, Hypotheses Generation without Feedback, human reviewers found the proposals scientifically valid but lacking in innovation. The reasoning behind material selection was minimal, and the focus on conventional materials and standard practices yielded low novelty.

When feedback loops were introduced, the hypotheses became more refined but continued to emphasize known materials with relatively shallow justifications. Although valid, the output remained constrained to well-documented methods, offering limited creativity.

In contrast, integrating both the Knowledge Graph and the critic feedback elicited the most positive responses. The system proposed a broader range of hypotheses and introduced new material combinations. Despite occasional oversimplifications and some lack of domain specificity, experts appreciated the heightened creativity and potential for innovation.

Overall, the human evaluations paralleled the automated results. The configuration without feedback consistently underperformed, the feedback-based system improved outcomes, and the system enhanced with the Knowledge Graph and feedback achieved the most favorable ratings. This consistency validates the automated evaluation with the evaluation Agent.

\section{Conclusion}

In this work, we introduced \textsc{\textbf{MatDesign}}, a novel benchmark for evaluating the ability of LLMs to generate innovative material hypotheses rather than merely reproducing pre-trained knowledge. Comprising 50 recently published journal papers (as of January 2024), it offers a reliable test bed for ensuring originality in the generated outputs.

Building on this benchmark, our \textsc{\textbf{AccelMat}} framework demonstrates promising results in producing both novel and feasible material hypotheses. Although the generated suggestions often fall short of providing the depth of reasoning or methodological details required for immediate practical application, they serve as a powerful starting point, enabling researchers to refine and extend the hypotheses further.

Finally, we show that our evaluation metric, though not on par with human-level analyses, aligns well with expert judgments. This corroborates the metric's reliability for hypothesis assessment in materials discovery and design and a feedback system which underscores its potential to guide iterative improvements in automated hypothesis-generation systems.








\section*{Ethics Statement}
We use AI assistants, specifically Grammarly and ChatGPT, were utilized to correct grammatical errors and restructure sentences.

\section*{Limitation}
While the curated dataset plays a crucial role in ensuring the novelty of generated hypotheses, its size (50 papers) may not fully capture the diversity of materials science research. Despite the experts' best efforts to identify all relevant papers published in renowned journals since January 2024, expanding the dataset as new papers become available could enhance its variety and enable the generation of more innovative hypotheses, providing a more comprehensive evaluation of the LLMs capabilities.

Another limitation arises from relying solely on LLMs for feedback or critique, as even unanimous agreement among state-of-the-art LLM-based critics does not guarantee scientific accuracy. The risk of hallucinated or flawed suggestions remains a challenge. Additionally, our LLM-based agents depend on human-provided constraints to guide hypothesis generation. Advancing the ability of LLMs to autonomously identify and apply relevant constraints could move us closer to achieving fully autonomous materials design and discovery.

\section*{Acknowledgement}
We thank the anonymous reviewers for their constructive suggestions. We extend our gratitude to the Research Computing (RC), and Enterprise Technology at Arizona State University for providing computing resources, and access to the ChatGPT enterprise version for experiments.

This research was supported by the Engineering Research and Development Center - Information Technology Laboratory (ERDC-ITL) under Contract No. W912HZ24C0022. Any opinions, findings and conclusions or recommendations expressed in this work are those of the author(s) and do not necessarily reflect the views of the ERDC-ITL.

\bibliography{references}

\appendix





\onecolumn
\appendix
\section*{Appendix}

\section{Open Source Experiments}

\begin{figure*}[h!]
    \centering
    \includegraphics[width=\textwidth]{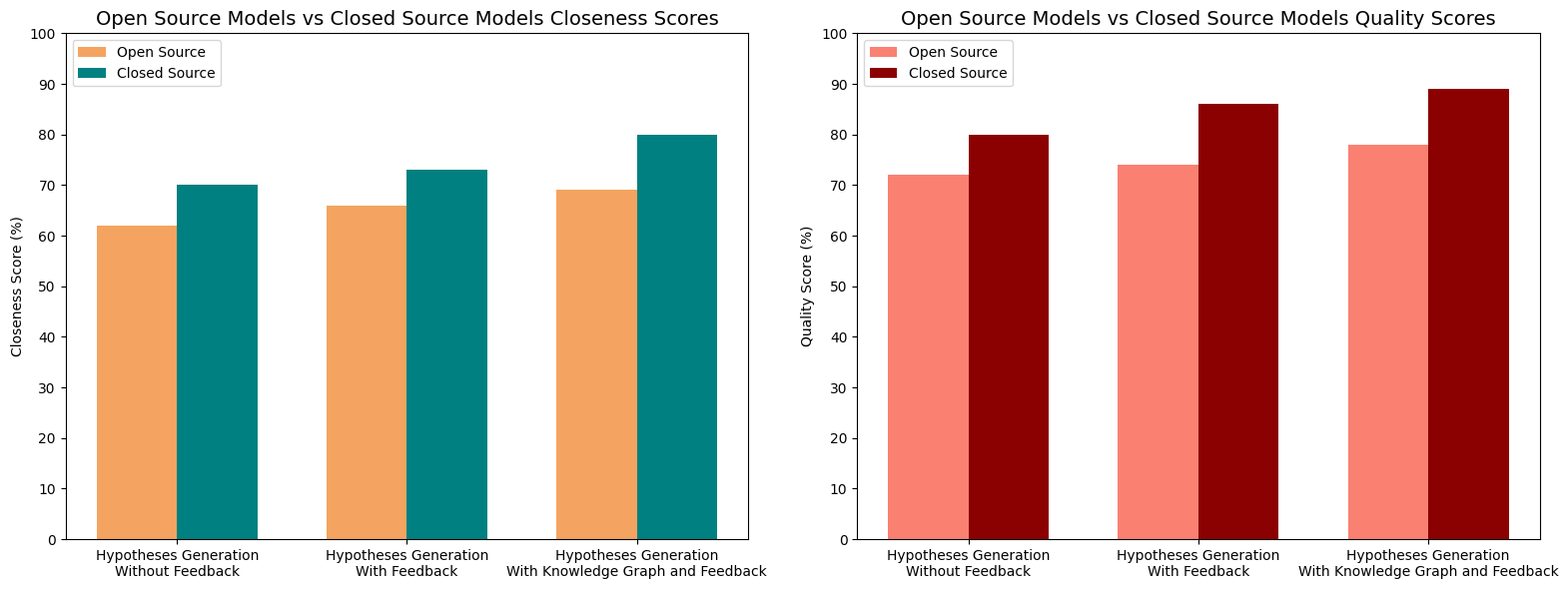}
    \caption{The left plot compares the performance of open-source and closed-source models on the Closeness metric, while the right plot compares their Quality scores. Both plots clearly show that closed-source models outperform open-source models significantly.}
    \label{closed source vs opensource comparison}
\end{figure*}

For comparing open and closed source models, we implemented three configurations described in Section \ref{Experiemnts} using open source models. Our setup included LLaMA-3.1 70B-Instruct as the Hypotheses Generator, while Gemma-2-27B-Instruct \citet{team2024gemma}, LLaMA-3.1 70B-Instruct, and Mixtral-8x22B-Instruct-v0.1 \citet{mistral_mixtral_8x22b} served as the three Critic Agents. LLaMA-3.1 70B-Instruct was employed as the Summarizer Agent, and OpenAI-o1-preview was used as the Evaluation Agent. As shown in Figure \ref{closed source vs opensource comparison}, closed source models consistently outperformed their open source counterparts in both closeness and quality metrics.

\label{appendix:open-source}

\section{Table of the six quality assessment criteria}
\label{section: table-six-criteria}

Table \ref{table: six-criteria-for-eval} summarizes the six quality assessment criteria defined in Section \ref{section:evaluation}.

\begin{table*}[ht]
\centering
\renewcommand{\arraystretch}{1.5} 
\resizebox{\textwidth}{!}{
\begin{tabular}{p{3cm}|p{3cm}|p{3cm}|p{3cm}|p{3cm}|p{3cm}} 
\toprule
\textbf{Evaluation Metric} & \textbf{1} & \textbf{2} & \textbf{3} & \textbf{4} & \textbf{5} \\
\hline
\textbf{Alignment with Research Objectives and Constraints} & Misaligned: Hypothesis does not relate to objectives and violates constraints. & Slightly Aligned: Hypothesis touches on the topic but misses key objectives or violates some constraints. & Moderately Aligned: Addresses some objectives but lacks focus on critical elements; meets most constraints. & Highly Aligned: Addresses most objectives and complies with major constraints. & Fully Aligned: Directly addresses all key objectives and fully adheres to all constraints. \\
\midrule
\textbf{Scientific Plausibility} & Not Plausible: Contradicts fundamental scientific principles. & Slightly Plausible: Contains significant inaccuracies or unsupported assumptions. & Moderately Plausible: Scientifically sound but includes minor inaccuracies or speculative elements. & Highly Plausible: Scientifically accurate with negligible issues. & Completely Plausible: Fully consistent with established knowledge and well-supported by theory. \\
\midrule
\textbf{Innovation and Novelty} & Not Innovative: Restates existing knowledge without introducing new ideas. & Slightly Innovative: Provides minimal new insights or slight variations on known concepts. & Moderately Innovative: Offers some original ideas or novel combinations of existing concepts. & Highly Innovative: Introduces significant new ideas or approaches. & Exceptionally Innovative: Groundbreaking ideas or methodologies that could revolutionize the field. \\
\midrule
\textbf{Testability} & Not Testable: Cannot be tested with current techniques. & Difficult to Test: Requires specialized or prohibitively expensive equipment. & Moderately Testable: Can be tested with available techniques but requires complex procedures. & Easily Testable: Testable using common equipment and straightforward procedures. & Highly Testable: Rapid, cost-effective validation with readily available resources and techniques. \\
\midrule
\textbf{Feasibility and Scalability} & Not Feasible: Cannot be implemented with current technology or resources. & Slightly Feasible: Theoretically possible but requires breakthroughs or high investment. & Moderately Feasible: Can be implemented on a small scale, but larger-scale production faces challenges. & Highly Feasible: Practical for small and moderate scales with manageable adjustments for scale-up. & Completely Feasible: Easily implemented and scaled from lab to industry with minimal changes or investment. \\
\midrule
\textbf{Impact Potential} & No Impact: Unlikely to contribute meaningfully to the field. & Low Impact: Offers minor contributions but lacks significant influence. & Moderate Impact: Can contribute positively in specific areas. & High Impact: Likely to drive significant advancements or address important challenges. & Transformative Impact: Capable of revolutionizing the field or providing solutions to major problems. \\
\bottomrule
\end{tabular}
}
\caption{Evaluation metrics and scales for assessing generated hypotheses based on alignment with research objectives, scientific plausibility, innovation, testability, feasibility, and impact potential.}
\label{table: six-criteria-for-eval}
\end{table*}

\section{Input Prompts}
\subsection{Goals and Constraints}

\begin{tcolorbox}[colback=black!5, colframe=black, title=Goal and constraints provided as input]
Goal:
\newline
Develop a scalable extrinsic self-healing coating system for corrosion protection of metallic structures in offshore environments. 
\newline\newline
Constraints:
\newline
1) The material should incorporate a self-healing mechanism triggered by a simple environmental factor (e.g., water).
\newline
2) The self-healing material should allow multiple healing events.
\newline
\end{tcolorbox}
\begin{tcolorbox}
3) The coating must maintain its structural integrity and protective capabilities even after mechanical damage.
\newline
4) The material should be compatible with scalable application techniques.
\newline
5) The healing mechanism should not rely on complex multi-component reactions but rather a single-component system
\end{tcolorbox}
\subsection{Prompt for experts list finder}

We use the below prompt for dynamically setting the behaviour of Hypotheses Generator relevant to the provided Goal. The output of this prompt is set as the system prompt for the Hypotheses Generator as shown in Appendix section \ref{box: hypothesis-generation-without-feedback} 
\begin{tcolorbox}[colback=black!5, colframe=black, title=Prompt for experts list finder]

Generate a list of experts required to achieve the below mentioned goal:\newline
\textbf{Develop a scalable extrinsic self-healing coating system for corrosion protection of metallic structures in offshore environments. .}\newline\newline Just list the top 5 experts in the format "Expert\_1, Expert\_2, Expert\_3, Expert\_4, Expert\_5"
\end{tcolorbox}

\subsection{Prompt for Hypotheses Generator}

\begin{tcolorbox}[colback=black!5, colframe=black, title= Prompt for Hypotheses Generator]
Generate 20 novel and innovative suggestions that strictly satisfy the provided goal and constraints. While building on existing scientific knowledge, also incorporate creative and innovative thinking to propose unique solutions that push the boundaries of current approaches.\newline
Ensure that each suggestion is scientifically plausible, grounded in reasoning, and offers a fresh perspective without introducing unsupported information. The ideas should have real-world potential and meet all constraints provided.
Provide reason for each suggestion. The suggestions must be in the below mentioned format in a JSON object. For example:\newline
\{\newline   
Suggestion\_1:\newline\{\newline
    Materials: "Give a comprehensive and detailed description of list of materials, along with their names and properties which achieve help the goal and satisfy all constraints"
    Methods: "Give detailed description of novel, innovative methods or techniques needed to develop the materials suggested"\newline
\end{tcolorbox}
\begin{tcolorbox}
    Reasoning: "Give a detailed reasoning for the suggestion"\newline\},\newline
    \vdots\newline
Suggestion\_20:\newline\{\newline
    Materials: "Give a comprehensive and detailed description of list of materials, along with their names and properties which achieve help the goal and satisfy all constraints"
    Methods: "Give detailed description of novel, innovative methods or techniques needed to develop the materials suggested"\newline
    Reasoning:"Give a detailed reasoning for the suggestion\newline\}\newline
    \}
\end{tcolorbox}
\label{box: hypothesis-generation-agent}

\subsection{Prompt for Critic Feedback}
\label{box: critic-agent}
\begin{tcolorbox}[colback=black!5, colframe=black, title= Prompt for Critic Feedback]
Goal\_statement:-\newline
\{goal\_statement\}\newline\newline
Constraints:-\newline\{constraint\_list\}\newline\newline
Suggestions:\newline{generated\_suggestions}.\newline\newline
Given the goal statement, constraints, and suggestions for materials design and discovery, evaluate each suggestion carefully. Provide detailed, constructive feedback on whether each suggestion has the potential to meet the goal and satisfy the constraints.\newline\newline
If any suggestion falls short, explain precisely where it fails and suggest specific improvements or adjustments that will help guide the generation agent to produce more accurate and innovative suggestions in the next cycle.\newline\newline
Ensure the feedback is actionable and focused on refining the generation process to align more closely with the goal and constraints in future iterations.\newline\newline
The detailed feedback should be in the below JSON format strictly:\newline\newline
        \{\newline"Feedback\_for\_suggestion\_1": \newline\{\newline
            "Meets\_the\_goal\_statement\_and\_satisfies\_all\_constraints\_strictly": "YES/NO",\newline
            "Reasoning": " "
        \newline\}\newline
        \vdots\newline
        "Feedback\_for\_suggestion\_20": \newline\{\newline
\end{tcolorbox}
\begin{tcolorbox}
            "Meets\_the\_goal\_statement\_and\_satisfies\_all\_constraints\_strictly": "YES/NO",\newline
            "Reasoning": " "
        \newline\},\newline
        "Overall\_Feedback\_for\_improvement\_for\_future\_suggestion\_generation": (list of points in a single line)\newline
        \newline\}
\end{tcolorbox}

\subsection{Prompt for Summarizer}
\label{box: summarizer-agent}
\begin{tcolorbox}[colback=black!5, colframe=black, title= Prompt for Summarizer]
Your task is to summarize feedback provided by multiple critics for a list of generated hypotheses. The summary must capture essential points and detailed insights of the feedback generated by all three critics. The final summary generated must be such that it will help in refining future generation of hypotheses. The feedback is as follows:\newline\newline
Critic\_1 feedback: \newline\{feedback\_from\_claude\}\newline\newline
Critic\_2 feedback:\newline \{feedback\_from\_gemini\}\newline\newline
Critic\_3 feedback: \newline\{feedback\_from\_gpt\}
\end{tcolorbox}

\section{Implemented Techniques }

\subsection{Hypotheses Generation Without feedback}
\label{box: hypothesis-generation-without-feedback}
The expert names are generated by the prompting Hypotheses Generator the experts list finder prompt, which it uses as system prompt for itself.
\begin{tcolorbox}[colback=black!5, colframe=black, title= System prompt for Hypotheses Generator ]
You are an innovative Materials Scientist, Corrosion Engineer, Polymer Chemist, Environmental Scientist, Chemical Engineer capable of doing impactful materials discovery and design. \newline

\end{tcolorbox}

\begin{tcolorbox}[colback=black!5, colframe=black, title= User input prompt to the Hypotheses Generator ]
Goal:
\newline
Develop a scalable extrinsic self-healing coating system for corrosion protection of metallic structures in offshore environments.
\end{tcolorbox}
\begin{tcolorbox}
Constraints:
\newline
1) The material should incorporate a self-healing mechanism triggered by a simple environmental factor (e.g., water)
\newline
2) The self-healing material should allow multiple healing events.
\newline
3) The coating must maintain its structural integrity and protective capabilities even after mechanical damage.
\newline
4) The material should be compatible with scalable application techniques.
\newline
5) The healing mechanism should not rely on complex multi-component reactions but rather a single-component system.

Generate 20 novel and innovative suggestions that strictly satisfy the provided goal and constraints. While building on existing scientific knowledge, also incorporate creative and innovative thinking to propose unique solutions that push the boundaries of current approaches.\newline
Ensure that each suggestion is scientifically plausible, grounded in reasoning, and offers a fresh perspective without introducing unsupported information. The ideas should have real-world potential and meet all constraints provided.
Provide reason for each suggestion. The suggestions must be in the below mentioned format in a JSON object. For example:\newline
\{\newline   
Suggestion\_1:\newline\{\newline
    Materials: "Give a comprehensive and detailed description of list of materials, along with their names and properties which achieve help the goal and satisfy all constraints"
    Methods: "Give detailed description of novel, innovative methods or techniques needed to develop the materials suggested"\newline
    Reasoning: "Give a detailed reasoning for the suggestion"\newline\},\newline
    \vdots\newline
Suggestion\_20:\newline\{\newline
    Materials: "Give a comprehensive and detailed description of list of materials, along with their names and properties which achieve help the goal and satisfy all constraints"
    Methods: "Give detailed description of novel, innovative methods or techniques needed to develop the materials suggested"\newline
    Reasoning:"Give a detailed reasoning for the suggestion\newline\}\newline
    \}

\end{tcolorbox}

\begin{tcolorbox}[colback=black!5, colframe=black, title=20 Hypotheses Generated by Hypotheses Generator]
Suggestion 1:\newline
Materials:Polyurea-based elastomer with embedded microtubules filled with a water-reactive monomer like cyanoacrylate. The microtubules are made from a biodegradable polymer such as polylactic acid (PLA).\newline
Methods:Use a co-extrusion process to create microtubules and embed them in the polyurea matrix. After embedding, the PLA can be dissolved to leave behind the microtubules. The cyanoacrylate can then be injected into these tubules. The final coating can be applied using standard spray-painting techniques.\newline
\end{tcolorbox}
\begin{tcolorbox}
Reasoning: Polyurea provides excellent mechanical properties and durability, while cyanoacrylate polymerizes rapidly upon contact with water. The use of PLA microtubules ensures a continuous supply of the healing agent to damaged areas, allowing for multiple healing events.\newline
\vdots\newline
Suggestion 20:\newline
Materials:Hybrid polymer matrix incorporating both hydrophilic and hydrophobic segments, with embedded nanocapsules containing a water-reactive monomer like methyl methacrylate (MMA).\newline
Methods:Disperse nanocapsules in the hybrid polymer matrix using mechanical mixing. Apply the coating using spray-painting techniques.\newline
Reasoning:MMA polymerizes upon contact with water, providing a quick and effective healing mechanism. The hybrid polymer matrix ensures compatibility with spray-painting and maintains the coating's mechanical integrity.\newline

\end{tcolorbox}

\begin{tcolorbox}[colback=black!5, colframe=black, title= Critic 1 GPT-4o feedback]
Feedback for suggestion 1: \newline
Meets the goal statement and satisfies all constraints strictly:YES. \newline
Reasoning: Polyurea-based elastomer with embedded microtubules filled with cyanoacrylate offers a scalable and robust self-healing mechanism triggered by water. The co-extrusion process ensures a continuous supply of the healing agent, and the application is compatible with spray-painting.\newline
\vdots \newline
Feedback for suggestion 20:\newline
Meets the goal statement and satisfies all constraints strictly:NO. \newline
Reasoning: Hybrid polymer matrices and nanocapsules introduce formulation complexity. Ensuring uniform nanocapsule distribution and multiple healing events in a scalable manner is challenging.\newline

Overall Feedback for future suggestion improvement: \newline
Focus on single-component healing agents activated by water.\newline
Ensure multiple healing events through vascular-like structures or microtubules\newline
Avoid complex multi-step synthesis processes.\newline
Ensure compatibility with scalable application techniques like spray-painting\newline
Address uniform distribution of self-healing agents within the polymer matrix.\newline
\end{tcolorbox}

\begin{tcolorbox}[colback=black!5, colframe=black, title= Critic 2 Claude 3.5 Sonnet feedback]
Feedback for suggestion 1:\newline
Meets the goal statement and satisfies all constraints strictly:YES.\newline
Reasoning: This suggestion meets all constraints. It uses water as a trigger, allows for multiple healing events through microtubules, maintains structural integrity, is compatible with spray-painting, and uses a single-component healing system.
\vdots \newline
Feedback for suggestion 20:\newline
\end{tcolorbox}
\begin{tcolorbox}
Meets the goal statement and satisfies all constraints strictly:NO.\newline
Reasoning: This suggestion fails to meet constraint 2 as it does not provide a mechanism for multiple healing events or a vascular-like structure for continuous supply of the healing agent.\newline

Overall Feedback for future suggestion improvement: \newline
Focus on developing vascular-like structures or continuous supply mechanisms to allow for multiple healing events \newline
Ensure all suggestions explicitly address each constraint, particularly the requirement for multiple healing cycles \newline
Explore innovative ways to integrate vascular structures with spray-painting techniques for large-scale application \newline
Consider combining successful elements from different suggestions, such as microtubules or vascular networks 
with various healing agents\newline
Investigate methods to create self-replenishing systems within the coating that are compatible with spray application\newline

\end{tcolorbox}

\begin{tcolorbox}[colback=black!5, colframe=black, title= Critic 3 Gemini 1.5 flash feedback]
Feedback for suggestion 1:\newline
Meets the goal statement and satisfies all constraints strictly:YES.\newline
Reasoning:The suggestion effectively addresses the goal and constraints. The use of polyurea for durability, cyanoacrylate for rapid healing, and PLA microtubules for continuous healing agent supply all contribute to a robust self-healing system. The co-extrusion method ensures scalability, and the spray-painting application aligns with industrial requirements.\newline
\vdots \newline
Feedback for suggestion 20:\newline
Meets the goal statement and satisfies all constraints strictly:YES.\newline
Reasoning: The polyester coatings with embedded microcapsules containing water-activated DMPA offer good mechanical properties and chemical resistance for offshore applications. The spray-painting compatibility and the single-component healing agent make this suggestion a strong candidate.\newline

Overall Feedback for future suggestion improvement: \newline
Explore a broader range of materials, consider incorporating bio-inspired self-healing mechanisms, and investigate the use of advanced characterization techniques to validate the proposed solutions.

\end{tcolorbox}

\begin{tcolorbox}[colback=black!5, colframe=black, title=Finalised Suggestions Agreed by all Critics]
Suggestion 1: \newline
Materials:Polyurea-based elastomer with embedded microtubules filled with a water-reactive monomer like cyanoacrylate. The microtubules are made from a biodegradable polymer such as polylactic acid (PLA).\newline
Methods:Use a co-extrusion process to create microtubules and embed them in the polyurea matrix. After embedding, the PLA can be dissolved to leave behind the microtubules. The cyanoacrylate can then be injected into these tubules. The final coating can be applied using standard spray-painting techniques.\newline
\end{tcolorbox}
\begin{tcolorbox}
Reasoning:Polyurea provides excellent mechanical properties and durability, while cyanoacrylate polymerizes rapidly upon contact with water. The use of PLA microtubules ensures a continuous supply of the healing agent to damaged areas, allowing for multiple healing events.\newline

Suggestion 17:\newline
Materials:Polyurethane-based elastomer with embedded microtubules filled with a water-reactive monomer like cyanoacrylate. The microtubules are made from a biodegradable polymer such as polylactic acid (PLA).\newline
Methods:Use a co-extrusion process to create microtubules and embed them in the polyurethane matrix. After embedding, the PLA can be dissolved to leave behind the microtubules. The cyanoacrylate can then be injected into these tubules. The final coating can be applied using standard spray-painting techniques.\newline
Reasoning:Polyurethane provides excellent mechanical properties and durability, while cyanoacrylate polymerizes rapidly upon contact with water. The use of PLA microtubules ensures a continuous supply of the healing agent to damaged areas, allowing for multiple healing events.\newline

Suggestion 9:\newline
Materials:Polyurethane-based elastomer with embedded microtubules filled with a water-reactive monomer like cyanoacrylate. The microtubules are made from a biodegradable polymer such as polylactic acid (PLA).\newline
Methods:Use a co-extrusion process to create microtubules and embed them in the polyurethane matrix. After embedding, the PLA can be dissolved to leave behind the microtubules. The cyanoacrylate can then be injected into these tubules. The final coating can be applied using standard spray-painting techniques.\newline
Reasoning:Polyurethane provides excellent mechanical properties and durability, while cyanoacrylate polymerizes rapidly upon contact with water. The use of PLA microtubules ensures a continuous supply of the healing agent to damaged areas, allowing for multiple healing events.\newline

\end{tcolorbox}

\subsection{Hypotheses Generation With feedback from Critics}
\label{box: hypothesis-generation-with-feedback-from-critics}

The input prompt to the Hypotheses Generator and Critics is as shown in the Appendix \ref{box: hypothesis-generation-without-feedback}.

\begin{tcolorbox}[colback=black!5, colframe=black, title= Summarised feedback from Summariser fed to the Hypotheses Generator for refinement]
Summary of Feedback from Multiple Critics on Generated Hypotheses: \newline
Common Approvals and Rejections:
Approval Trends:
Polyurethane matrix with cyanoacrylate and  epoxy resin with TMPTA were universally approved for their scalability, effectiveness in multiple healing events, and compatibility with industrial applications like spray-painting.
Fluoropolymer coatings with n-butyl acrylate-filled channels** and **polyurea-based elastomer with polyisocyanate-filled channels were consistently approved for their chemical resistance and suitability for harsh environments, as well as their scalable application methods.\newline
\vdots\newline
Single-component healing systems triggered by water were highly favored across all suggestions.
For future hypothesis generation, focus on combining simple, scalable manufacturing processes with innovative material choices and healing mechanisms. Prioritize single-component, water-activated healing agents compatible with industrial application techniques. Incorporate advanced characterization methods and additional functionalities to refine and validate the solutions robustly. Exploring hybrid systems and alternative stimuli can further enhance the effectiveness and adaptability of self-healing coatings.
\end{tcolorbox}

\begin{tcolorbox}[colback=black!5, colframe=black, title= Refined Hypotheses Generated by Hypotheses Generator by incoprorating feedback]
Suggestion 1: \newline
Materials:Polyurethane matrix with cyanoacrylate-filled vascular-like channels. \newline
Methods:3D print a network of channels within the polyurethane matrix using a sacrificial material like PVA. Dissolve the PVA post-formation to leave hollow channels and inject cyanoacrylate. Apply the final coating using spray-painting techniques.\newline
Reasoning:Polyurethane provides durability while cyanoacrylate offers a rapid water-activated healing mechanism. The vascular-like structure allows multiple healing events, and the method is compatible with scalable techniques like spray-painting.\newline
\vdots \newline
Suggestion 20:\newline
Materials:Polyurethane matrix with embedded microcapsules containing isocyanate.\newline
Methods:Disperse microcapsules containing isocyanate uniformly in a polyurethane matrix. Apply the coating using spray-painting techniques. Upon damage, isocyanate reacts with water to form a protective polymer.\newline
Reasoning:Isocyanate offers a simple and efficient water-activated healing mechanism. Polyurethane provides durability, and microcapsules ensure even distribution of the healing agent.\newline
\end{tcolorbox}

\begin{tcolorbox}[colback=black!5, colframe=black, title= Critic 1 GPT-4o feedback]
Feedback for suggestion 1: \newline
Meets the goal statement and satisfies all constraints strictly:YES. \newline
Reasoning: Polyurethane matrix with cyanoacrylate-filled vascular-like channels is durable, has a rapid water-activated healing mechanism, allows multiple healing events, and is compatible with scalable techniques like spray-painting. \newline
\vdots \newline
Feedback for suggestion 20:\newline
Meets the goal statement and satisfies all constraints strictly:YES. \newline
Reasoning: Polyurethane matrix with isocyanate-filled microcapsules ensures a simple, efficient water-activated healing mechanism, durability, and compatibility with spray-painting techniques.
.\newline

Overall Feedback for future suggestion improvement: \newline
All of the suggestions satisfy the constraints and align with the goals
\end{tcolorbox}

\begin{tcolorbox}[colback=black!5, colframe=black, title= Critic 2 Claude 3.5 Sonnet feedback]
Feedback for suggestion 1:\newline
Meets the goal statement and satisfies all constraints strictly:YES.\newline
Reasoning:This suggestion meets all constraints. It uses water as a trigger for cyanoacrylate, allows for multiple healing events through vascular-like channels, maintains structural integrity with polyurethane, is compatible with spray-painting, and uses a single-component healing system.
\newline
\vdots \newline
Feedback for suggestion 20:\newline
\end{tcolorbox}
\begin{tcolorbox}
Meets the goal statement and satisfies all constraints strictly:YES.\newline
Reasoning: This suggestion satisfies all constraints. It uses water as a trigger for isocyanate, provides multiple healing events through microcapsules, maintains structural integrity with polyurethane, is compatible with spray-painting, and uses a single-component healing system..\newline

Overall Feedback for future suggestion improvement: \newline
I agree with all of the suggestion, but they can be further improved :\newline
Explore novel combinations of matrix materials and healing agents to enhance performance and durability ...

\end{tcolorbox}

\begin{tcolorbox}[colback=black!5, colframe=black, title= Critic 3 Gemini 1.5 flash feedback]
Feedback for suggestion 1:\newline
Meets the goal statement and satisfies all constraints strictly:YES.\newline
Reasoning:This suggestion effectively addresses the goal and constraints. The use of polyurethane for durability, cyanoacrylate for rapid healing, and PVA microtubules for continuous healing agent delivery all contribute to a robust self-healing system. The 3D printing method ensures scalability, and the spray-painting application aligns with industrial requirements.\newline
\vdots \newline
Feedback for suggestion 20:\newline
Meets the goal statement and satisfies all constraints strictly:YES.\newline
Reasoning:Fluoropolymers offer chemical resistance, and n-butyl acrylate provides a water-activated healing mechanism. The method is scalable and effective.\newline

Overall Feedback for future suggestion improvement: \newline
All suggestions align with constrainst and goals, it can be further improved .....

\end{tcolorbox}

\subsection{Hypotheses Generation With Knowledge Graph and Feedback from Critics}
\label{box: hypothesis-generation-with-KG-and-feedback}

\begin{tcolorbox}[colback=black!5, colframe=black, title= User input prompt to the Hypotheses Generator with Knowledge Graphs]
Goal:
\newline
Develop a scalable extrinsic self-healing coating system for corrosion protection of metallic structures in offshore environments. 
\newline\newline
Constraints:
\newline
1) The material should incorporate a self-healing mechanism triggered by a simple environmental factor (e.g., water).
\newline
2) The self-healing material should allow multiple healing events.
\newline
3) The coating must maintain its structural integrity and protective capabilities even after mechanical damage.
\newline
\end{tcolorbox}
\begin{tcolorbox}
4) The material should be compatible with scalable application techniques.
\newline
5) The healing mechanism should not rely on complex multi-component reactions but rather a single-component system.

Relevant information from trusted Knowledge Graphs:\newline

Suggested Materials: \newline
Aluminum: Aluminum is known for its corrosion resistance and is commonly used in offshore environments.\newline
Graphene: Graphene has excellent barrier properties and mechanical strength, making it suitable for self-healing coatings.\newline
Epoxy: Epoxy is widely used in coatings for its strong adhesion and durability, essential for corrosion protection.\newline
Zinc: Zinc is often used in coatings for its sacrificial protection properties, which are beneficial for corrosion resistance.\newline

Suggested Properties:\newline
Corrosion Resistance: Essential for protecting metallic structures in corrosive offshore environments.\newline
Barrier Properties: Important for preventing the ingress of corrosive agents like water and salts.\newline
Mechanical Property: Necessary to ensure the coating can withstand mechanical damage and maintain its integrity.\newline
Adhesions: Critical for ensuring the coating remains attached to the substrate, especially in harsh conditions.\newline

TASK: \newline
Generate 20 novel and innovative suggestions that strictly satisfy the provided goal and constraints. Use the information provided from the trusted knowledge graph as a reference to generate suggestions, but aim to propose novel combinations or entirely new materials and methods. While building on existing scientific knowledge, also incorporate creative and innovative thinking to propose unique solutions that push the boundaries of current approaches. \newline
Ensure that each suggestion is scientifically plausible, grounded in reasoning, and offers a fresh perspective. The ideas should have real-world potential and meet all constraints and the goal provided. Provide scientific reasoning for each suggestion.The suggestions must be in the below mentioned format in a JSON object For example:\newline
\{\newline   
Suggestion\_1:\newline\{\newline
    Materials: "Give a comprehensive and detailed description of list of materials, along with their names and properties which achieve help the goal and satisfy all constraints"
    Methods: "Give detailed description of novel, innovative methods or techniques needed to develop the materials suggested"\newline
    Reasoning: "Give a detailed reasoning for the suggestion"\newline\},\newline
    \vdots\newline
Suggestion\_20:\newline\{\newline
    Materials: "Give a comprehensive and detailed description of list of materials, along with their names and properties which achieve help the goal and satisfy all constraints"
\end{tcolorbox}
\begin{tcolorbox}
    Methods: "Give detailed description of novel, innovative methods or techniques needed to develop the materials suggested"\newline
    Reasoning:"Give a detailed reasoning for the suggestion\newline\}\newline
    \}

\end{tcolorbox}

\begin{tcolorbox}[colback=black!5, colframe=black, title=20 Hypotheses Generated by Hypotheses Generator with Knowledge Graphs]
Suggestion 1:\newline
Materials:A single-component polyurethane matrix combined with zinc oxide nanoparticles and tannin-derived microcapsules. The tannin-derived microcapsules offer a simple, natural healing mechanism activated by water exposure, while zinc oxide provides dual protection and healing properties.\newline
Methods:Utilize a scalable spray-painting method to apply the polyurethane coating. The zinc oxide nanoparticles enhance the barrier properties, and the tannin-derived microcapsules release their content upon contact with water, initiating self-healing.\newline
Reasoning:This design leverages simplicity and sustainability by using tannin-derived agents, aligning with critics' feedback to avoid complex silane-based solutions. Zinc oxide offers excellent corrosion resistance, making it suitable for offshore environments.\newline

Suggestion 2:\newline
Materials:Epoxy matrix with embedded graphene oxide and water-responsive natural polymers, such as chitosan. Graphene oxide provides excellent barrier properties, while chitosan acts as a natural, biodegradable healing agent.\newline
Methods:Develop a sprayable epoxy coating where graphene oxide enhances mechanical strength. Chitosan microcapsules release upon water contact, enabling multiple healing cycles.\newline
Reasoning:Graphene oxide enhances both mechanical and barrier properties. The use of chitosan aligns with sustainability goals, offering a biodegradable option that critics favor.\newline
.\newline
.\newline
.\newline
Suggestion 20:\newline
Materials:Polysiloxane matrix with embedded graphene oxide and tannin-derived microcapsules.\newline
Methods:Create sprayable polysiloxane coatings where graphene oxide enhances barrier properties. Tannin-derived microcapsules release healing agents upon water exposure.\newline
Reasoning:This suggestion uses natural healing agents and graphene oxide for enhanced performance, aligning with simplicity and environmental considerations.
\end{tcolorbox}

\section{Evaluation Prompts for OpenAI-o1-preview}
\label{Evaluation Prompts for OpenAI-o1-preview}
\begin{tcolorbox}[colback=black!5, colframe=black, title= Prompt for Closeness Measurement]
You will be provided with a number of suggested hypotheses which you have to rate by comparing them with the ground truth and based on your rating you have to rank from most rated to lowest rated. The ground-truth has been provided through real-world scientific experimentation to achieve the goal statement below. For the rating and ranking you have to provide detailed reasoning. You have to rank each suggested hypothesis on an individual basis; do not club the ratings of multiple hypotheses together.  Below are the rules for rating and ranking.
\newline
Rules: \newline
1) Concept Overlap Scale:
\newline
Definition: This scale evaluates how well the core ideas,
methods, and scientific concepts in the generated hypothesis
align with those in the ground truth hypothesis.
\newline 
Scale:
\newline
1 - No Overlap: The generated hypothesis contains entirely different concepts with no connection to the core ideas or methods in the ground truth hypothesis.
\newline
2 - Minimal Overlap: A few general or broad ideas may be similar, but most key concepts or methods are either missing or significantly different.
\newline
3 - Moderate Overlap: Several core ideas and methods overlap with the ground truth, but some critical concepts are either missing or misrepresented.
\newline
4 - High Overlap: Most of the core concepts and methods in the generated hypothesis match well with the ground truth, with only minor differences.
\newline
5 - Complete Overlap: The generated hypothesis fully mirrors the core ideas, methods, and scientific concepts in the ground truth hypothesis.
\newline
\newline
2) Property Overlap:
\newline
Definition: This scale assesses how closely the material properties described in the generated hypothesis align with those in the ground truth, including quantitative values and qualitative descriptions.
\newline
Scale:
\newline
1 - Not Similar: The generated hypothesis has no overlapping or similar properties to the ground truth hypothesis. Properties are either missing or completely different.
\newline
2 - Slightly Similar: A few properties may match, but most are different in magnitude or type. Qualitative descriptions may vary significantly.
\newline
3 - Moderately Similar: Some important properties are similar, though others differ in specific values or qualitative descriptions.
\newline
4 - Highly Similar: Most of the key properties align well with the ground truth, with only minor differences in values or descriptions.
\newline
5 - Perfect Match: The generated hypothesis fully matches the material properties of the ground truth, including both quantitative and qualitative aspects.
\newline
\newline
3) Keyword Matching:
\newline
Definition: This scale evaluates how well specific entities (e.g., material names, chemical compounds, synthesis methods) and keywords in the generated hypothesis match those in the ground truth.
\newline
Scale:
\newline
1 - No Match: None of the keywords or entities in the generated hypothesis match those in the ground truth. There is a complete divergence in terms of critical terms and concepts.
\newline
2 - Minimal Match: A small number of non-critical keywords or entities match, but most key terms are either missing or incorrect.
\end{tcolorbox}
\begin{tcolorbox} 
3 - Partial Match: Several important keywords and entities match, though some are either missing or not fully aligned.
\newline
4 - High Match: Most critical keywords and entities match, with only a few minor discrepancies.
\newline
5 - Complete Match: All key keywords and entities in the generated hypothesis align exactly with the ground truth.
\newline
\newline
The ground truth is:
\newline
Materials:
\newline
Core–shell nanofibers synthesized using coaxial electrospinning. Organosilane compounds, specifically silyl esters, used as the self-healing agent. Metallic substrates (e.g., steel) for corrosion testing.
\newline
Methods:
\newline
Coaxial electrospinning of core–shell nanofibers: The self-healing material was developed using coaxial electrospinning to create core–shell nanofibers with an organosilane compound (silyl ester) as the healing agent.
\newline
Overcoming scalability challenges with spray painting: A viable spray-painting technique to apply the nanofiber-based coating, making it scalable for large structures. Spray painting ability was achieved by prior dispersion of the nanofibres.
\newline
Incorporation of water-reactive organosilanes: Water-reactive organosilane (silyl ester) was incorporated as the healing agent within the core of the nanofibers, allowing the coating to heal upon exposure to water without requiring any additional catalysts.
\end{tcolorbox}

\begin{tcolorbox}[colback=black!5, colframe=black, title= Prompt for Quality Measurement]
You will be provided with a number of suggested hypotheses which you have to rate by assesing their quality based on 6 criterias defined below. And based on your rating you have to rank from most rated to lowest rated. For the rating and ranking you have to provide detailed reasoning. You have to rank each suggested hypothesis on an individual basis; do not club the ratings of multiple hypotheses together.  Below are the rules for rating and ranking.
\newline
Rules:
\newline
1) Alignment with Research Objectives and Constraints:
\newline
Definition: Assesses how directly and effectively the generated hypothesis addresses the objectives specified in the goal statement while adhering to all given constraints and incorporating provided key points or clues.
\newline
Scale:
\newline
1 - Misaligned: The hypothesis does not relate to the re-
search objectives in any meaningful way and ignores or
violates specified constraints.
\newline
2 - Slightly Aligned: The hypothesis touches on the general
topic but misses key objectives or questions. It adheres to
some constraints but overlooks or violates others.
\newline
3 - Moderately Aligned: The hypothesis addresses some
aspects of the research objectives but lacks focus on critical elements. It meets most constraints but may miss minor ones or incorporate them incorrectly.
\newline
4 - Highly Aligned: The hypothesis effectively addresses
most research objectives and questions. It complies with
all major constraints and key points, with only minor omis-
sions.
\newline
5 - Fully Aligned: The hypothesis directly and comprehensively addresses all key objectives and questions of the research goals. It completely adheres to all specified constraints and fully incorporates all key points and clues.
\newline
\newline
2) Scientific Plausibility:
\newline
Definition: Assesses whether the hypothesis is grounded in
established scientific principles and theories within material science.
\newline
Scale:
\newline
1 - Not Plausible: Contradicts fundamental scientific laws
or principles.
\newline
2 - Slightly Plausible: Contains significant scientific inaccuracies or unsupported assumptions.
\newline
3 - Moderately Plausible: Generally scientifically sound but
includes minor inaccuracies or speculative elements.
\newline
4 - Highly Plausible: Scientifically accurate and well-
supported by current theories with negligible issues.
\newline
5 - Completely Plausible: Fully consistent with established
scientific knowledge and principles, with strong theoretical
support.
\newline
\newline
3) Innovation and Novelty:
\newline
Definition: Measures the degree to which the hypothesis
introduces original ideas, perspectives, or methods not previously documented.
\newline
Scale:
\newline
1 - Not Innovative: Restates existing knowledge without
introducing new ideas.
\newline
2 - Slightly Innovative: Provides minimal new insights or
slight variations on known concepts.
\newline
3 - Moderately Innovative: Offers some original ideas or
novel combinations of existing concepts.
\newline
4 - Highly Innovative: Introduces significant new ideas or
approaches that could advance the field.
\newline
5 - Exceptionally Innovative: Presents groundbreaking ideas
or methodologies with the potential to revolutionize the
field
\end{tcolorbox}
\begin{tcolorbox}
4) Testability:
\newline
Definition: Evaluates how easily and effectively the hy-
pothesis can be tested experimentally, considering the avail-
ability of required techniques, equipment, and resources in
materials science.
\newline
Scale:
\newline
1 - Not Testable: The hypothesis cannot be tested with current or foreseeable experimental techniques.
\newline
2 - Difficult to Test: Testing requires highly specialized,
rare, or prohibitively expensive equipment and resources.
\newline
3 - Moderately Testable: Can be tested with available techniques, but requires complex procedures or significant resources.
\newline
4 - Easily Testable: Can be tested using common laboratory
equipment and straightforward procedures.
\newline
5 - Highly Testable: Allows for rapid, cost effective experimental validation with readily available resources and
techniques.
\newline
\newline
5) Feasibility and Scalability:
\newline
Definition: Evaluates the practicality of implementing the
hypothesis at various scales, from laboratory to industrial
applications, focusing on existing infrastructure, technology, cost, and effort required to scale the process.
\newline
Scale:
\newline
1 - Not Feasible: Cannot be implemented with current technology, materials, or resources, or faces insurmountable
obstacles in scale-up.
\newline
2 - Slightly Feasible: Implementation is theoretically possible but would require breakthroughs or prohibitively high investment in materials, equipment, or processes. Scaling up significantly increases the complexity and costs.
\newline
3 - Moderately Feasible: Can be implemented on a small
scale (e.g., lab setting), but larger-scale production faces
major challenges, such as cost, resource availability, or technological limitations.
\newline
4 - Highly Feasible: Practical to implement with current
technology, materials, and methods on both small and moderate scales. Scaling up to industrial production would
require some adjustments but is manageable without substantial redesign or new resources.
\newline
5 - Completely Feasible: Easily implemented and scaled
from lab to industrial scale with minimal changes or additional investment. Resources, techniques, and materials
are widely available, and the process is cost-effective at all scales.
\newline
\newline
6) Impact Potential:
\newline
Definition: Assesses the hypothesis’s potential to significantly advance the field or address critical challenges in material science.
\newline
Scale:
\newline
1 - No Impact: Unlikely to contribute meaningfully to the
field.
\newline
2 - Low Impact: May offer minor contributions but lacks
significant influence.
\newline
3 - Moderate Impact: Has the potential to contribute positively in specific areas.
\newline
4 - High Impact: Likely to drive significant advancements
or address important challenges.
\newline
5 - Transformative Impact: Capable of revolutionizing the
field or providing solutions to major problems
\end{tcolorbox}

\section{Evaluated Hypotheses by OpenAI-o1-preview}
\label{Evaluated Hypotheses by OpenAI-o1-preview}
\begin{tcolorbox}[colback=black!5, colframe=black, title= Hypotheses Evaluated with Closeness Metric]
Suggestion 6:
\newline
Materials: Incorporate a polyurea coating with dispersed microcapsules containing a water-activated healing agent such as a silane-based compound.
\newline
Methods: Utilize in-situ polymerization to encapsulate the healing agent within a polyurea matrix. Apply using high-pressure spray techniques suitable for polyurea coatings.
\newline
Reasoning: Polyurea provides excellent chemical and abrasion resistance, ideal for offshore environments. The water-activated silane-based healing agent allows for immediate response to damage, ensuring continuous protection.
\newline
\newline
Evaluation:
\newline
Concept Overlap (3 - Moderate Overlap): This suggestion shares several core ideas with the ground truth, such as using a water-activated silane-based healing agent and applying the coating through spray techniques. However, it differs significantly in the encapsulation method and the type of polymer matrix used. While the ground truth employs coaxial electrospinning to create core–shell nanofibers, this suggestion uses microcapsules dispersed in a polyurea matrix. The difference in the delivery system (nanofibers vs. microcapsules) marks a critical divergence in methods.
\newline
Property Overlap (3 - Moderately Similar): Both coatings aim for immediate self-healing upon exposure to water and provide corrosion protection. The use of a silane-based healing agent is a common property. However, the suggestion lacks specific quantitative properties like the 97.5\% recovery of protective properties mentioned in the ground truth. Additionally, the mechanical properties may differ due to the use of polyurea instead of nanofiber structures.
\newline
Keyword Matching (3 - Partial Match): The suggestion matches several important keywords such as "silane-based healing agent," "water-activated," and "spray application." However, it does not include critical terms from the ground truth like "coaxial electrospinning," "core–shell nanofibers," or "organosilane compound (silyl ester)."
\newline
\newline
Suggestion 7:
\newline
Materials: Design a coating using a poly(dimethylsiloxane) (PDMS) matrix with embedded microcapsules containing a water-sensitive alkoxy silane healing agent.
\newline
Methods: Integrate microencapsulation techniques to uniformly disperse capsules within the PDMS matrix. Use spray application for even distribution over large surfaces.
\newline
Reasoning: PDMS is known for its flexibility and hydrophobicity, providing a durable base. The alkoxy silane reacts with water to form a protective siloxane layer, ensuring multiple healing cycles and robust corrosion resistance.
\newline
\newline
Evaluation:
\newline
Concept Overlap (4 - High Overlap): This suggestion closely aligns with the ground truth by utilizing a silane-based healing agent that reacts with water to initiate self-healing. Both methods involve embedding the healing agent within a polymer matrix and applying the coating via spray techniques. The primary difference lies in the use of microcapsules within a PDMS matrix instead of core–shell nanofibers created by coaxial electrospinning.
\newline
Property Overlap (4 - Highly Similar): Both coatings are designed to self-heal upon exposure to water and offer corrosion resistance. The suggestion's use of PDMS provides flexibility and durability, which are advantageous properties similar to those achieved with nanofiber coatings. While specific quantitative recovery rates are not mentioned, the focus on multiple healing cycles and robust protection indicates a strong alignment with the ground truth properties.
\newline
\end{tcolorbox}
\begin{tcolorbox}
Keyword Matching (4 - High Match): The suggestion matches most critical keywords, including "silane healing agent," "water-sensitive," "spray application," "self-healing," and "corrosion resistance." Minor discrepancies include the absence of terms like "coaxial electrospinning" and "core–shell nanofibers," but the overall keyword alignment is strong.
\newline
\newline
Suggestion 1:
\newline
\vdots \newline
====================================================================
\newline
Ranking from Most to Least Aligned with the Ground Truth:
\newline
Suggestion 7
\newline
Reasoning: Suggestion 7 exhibits high overlap in concepts, properties, and keywords. It closely mirrors the ground truth's approach by using a silane-based healing agent activated by water, embedded within a polymer matrix, and applied via spray techniques. The minor differences do not significantly detract from the overall alignment.
\newline
\newline
Suggestion 6
\newline
Reasoning: Suggestion 6 shows moderate overlap in all categories. It shares the use of a silane-based healing agent and spray application methods but differs in the encapsulation technique and the polymer matrix used. The core concepts are somewhat aligned but lack the depth of similarity seen in Suggestion 7.
\newline
\newline
Suggestion 1
\newline
Reasoning: Suggestion 1 has minimal overlap with the ground truth. The healing agent and encapsulation methods are different, and there is a lack of critical keyword alignment. While it shares the broad idea of a water-activated self-healing coating, the significant differences in materials and methods place it lowest in the ranking.
\newline
\newline
Final Rankings:
\newline
Suggestion 7
\newline
Concept Overlap: 4
\newline
Property Overlap: 4
\newline
Keyword Matching: 4
\newline
\newline
Suggestion 6
\newline
Concept Overlap: 3
\newline
Property Overlap: 3
\newline
Keyword Matching: 3
\newline
\newline
Suggestion 1
\newline
Concept Overlap: 2
\newline
Property Overlap: 2
\newline
Keyword Matching: 2
\end{tcolorbox}

\begin{tcolorbox}[colback=black!5, colframe=black, title= Hypotheses Evaluated with Quality Metric]
Suggestion 1
\newline
Materials: Polyurethane matrix with cyanoacrylate-filled vascular-like channels.
\newline
Methods: 3D print a network of channels within the polyurethane matrix using a sacrificial material like PVA. Dissolve the PVA post-formation to leave hollow channels and inject cyanoacrylate. Apply the final coating using spray-painting techniques.
\newline
Reasoning: Polyurethane provides durability while cyanoacrylate offers a rapid water-activated healing mechanism. The vascular-like structure allows multiple healing events, and the method is compatible with scalable techniques like spray-painting.
\newline
\newline
Evaluation:
\newline
Alignment with Research Objectives and Constraints: Score: 3 (Moderately Aligned)
\newline
Reasoning: 
The hypothesis addresses the objective of creating a self-healing coating with water-activated mechanisms. However, incorporating vascular-like channels via 3D printing may not be fully compatible with spray-painting techniques, which is a specified constraint. The use of 3D printing and sacrificial materials could complicate scalability and industrial application.
\newline
\newline
Scientific Plausibility: Score: 4 (Highly Plausible)
\newline
Reasoning:
The use of polyurethane and cyanoacrylate is scientifically sound, and the concept of vascular networks for self-healing is established in materials science. Minor challenges may exist in integrating 3D printed channels with spray-painting, but overall, the hypothesis is plausible.
\newline
\newline
Innovation and Novelty: Score: 4 (Highly Innovative)
\newline
Reasoning:
Combining vascular networks with spray-paintable coatings introduces a novel approach to self-healing materials. The integration of 3D printing techniques in creating self-healing coatings is a significant advancement.
\newline
\newline
Testability: Score: 2 (Difficult to Test)
\newline
Reasoning:
Testing this hypothesis requires specialized equipment like 3D printers capable of creating vascular networks within polymers and handling sacrificial materials. This complexity makes experimental validation more challenging and resource-intensive.
\newline
\newline
Feasibility and Scalability: Score: 2 (Slightly Feasible)
\newline
Reasoning:
Implementing vascular networks within coatings on an industrial scale is challenging. The process involves complex steps that are not easily scalable with current spray-painting technologies. Scaling up would require significant adjustments and investment.
\newline
\newline
Impact Potential: Score: 4 (High Impact)
\newline
Reasoning:
If successfully implemented, this approach could significantly advance self-healing coatings by enabling multiple healing events and improving longevity, addressing critical challenges in corrosion protection.
\newline
\newline
Suggestion 2
\newline
Materials: Polyurea-based elastomer with polyisocyanate-filled vascular-like channels.
\newline
Methods: Use 3D printing to create a network of channels within the polyurea matrix using a sacrificial material. Dissolve the sacrificial material to leave the channels, then inject polyisocyanate. Apply using spray-painting techniques.
\newline
Reasoning: Polyurea and polyisocyanate are effective for corrosion protection. The vascular-like structure ensures multiple healing events, and the process is scalable.
\newline
\newline
\end{tcolorbox}
\begin{tcolorbox}
Evaluation:
\newline
Alignment with Research Objectives and Constraints: Score: 3 (Moderately Aligned)
\newline
\vdots\newline
===================================================================

Ranking from Highest to Lowest Rated
\newline
Based on the evaluations, the suggestions can be ranked according to their overall scores:
\newline
Suggestions 4, 5, 6, 7, 9, 11, 13, 15, 17, 18, 19, 20 (Tied for Highest Rating)
\newline
Reasoning: These suggestions scored consistently high across all criteria, fully aligning with research objectives and constraints, being scientifically plausible, highly testable, completely feasible, and having high impact potential.
\newline
\newline
Suggestions 10 and 16 (Next Highest Rating)
\newline
Reasoning: These scored slightly lower in testability and feasibility due to minor challenges with fluoropolymer processing but still rated highly overall.
\newline
\newline
Suggestions 1, 2, 3, 8, 12, 14 (Lower Rating)
\newline
Reasoning: These suggestions, while innovative, faced challenges in alignment due to the use of vascular-like channels which complicate compatibility with spray-painting techniques, resulting in lower scores for testability and feasibility.
\end{tcolorbox}

\section{Closeness Evaluation with BERTScore}
\label{Closeness Evaluation with BERTScore}
To evaluate the similarity between the generated hypotheses and the ground truth hypotheses, we computed BERTScore \citet{zhang2019bertscore} with the MaterialsBERT model \citep{shetty2023general}. The calculated scores for the proposed methods are as follows:
\begin{itemize}
    \item \textbf{Hypotheses Generation Without Feedback:} 52.66\%
    \item \textbf{Hypotheses Generation With Feedback:} 60.59\%
    \item \textbf{Hypotheses Generation With Knowledge Graph and Feedback:} 50.30\%
\end{itemize}

However, these scores do not align with the evaluations conducted by LLM-based methods or human experts in our study where Hypotheses Generation With Knowledge Graph and Feedback method achieves the best scores. Moreover, interpreting these scores within the context of our task poses challenges. For example, while there is a 7.93\% improvement between the first two methods, it is unclear how this translates into factors such as the alignment of the mentioned properties, or the similarity of the methods described. This lack of interpretability makes BERTScore less effective for meaningfully assessing the performance of our methods in this domain.

\end{document}